\RequirePackage{amsmath}
\documentclass[runningheads]{llncs}
\usepackage{color,array}
\usepackage{tabularx} 
\usepackage[section]{placeins}
\usepackage{afterpage}
\usepackage{algorithmic}

\usepackage{graphicx}
\usepackage{graphics}
\usepackage{subfigure}

\usepackage{bm}
\usepackage{latexsym}
\usepackage{booktabs}
\usepackage{threeparttable}
\usepackage{epsf}
\usepackage{algorithm}
\usepackage{color,colortbl}
\usepackage{babel}
\renewcommand{\algorithmicensure}{\textbf{Output:}}
\renewcommand{\algorithmicrequire}{\textbf{Input:}}
\usepackage{hyperref}
\hypersetup{
	colorlinks=true,
	linkcolor=blue,
	anchorcolor=blue,
	citecolor=blue}

\usepackage{multirow}
\usepackage[normalem]{ulem}
\useunder{\uline}{\ul}{}
\graphicspath{{img/},{pics/}}
\usepackage[T1]{fontenc}
\usepackage{amsmath}
\usepackage{subfigure}
\usepackage{color}
\usepackage[misc]{ifsym}

\usepackage{hyphenat}
\begin{document}
\title{Logits Poisoning Attack in Federated Distillation}
%
%

\author{
Yuhan Tang\inst{1,2} \and
Zhiyuan Wu\inst{3,4} \and
Bo Gao\inst{1,2} \textsuperscript{(\Letter)} \and
Tian Wen\inst{3} \and \\
Yuwei Wang\inst{3} \and Sheng Sun\inst{3}
}
\authorrunning{Y. Tang et al.}
\institute{
Engineering Research Center of Network Management Technology for High-Speed Railway of Ministry of Education, School of Computer and Information Technology, Bejing Jiaotong University, Beijing, China 
\and Collaborative Innovation Center of Railway Traffic Safety, Beijing Jiaotong University, Beijing, China
\and
Institute of Computing Technology, Chinese Academy of Sciences, Beijing, China \and
University of Chinese Academy of Sciences, Beijing, China
\\
\email{
20722020@bjtu.edu.cn \quad wuzhiyuan22s@ict.ac.cn \quad bogao@bjtu.edu.cn\\
marrowd611@gmail.com \quad \{sunsheng,ywwang\}@ict.ac.cn}
}

%
\maketitle              
%

\begin{abstract}
Federated Distillation (FD) is a novel and promising distributed machine learning paradigm, where knowledge distillation is leveraged to facilitate a more efficient and flexible cross-device knowledge transfer in federated learning. By optimizing local models with knowledge distillation, FD circumvents the necessity of uploading large-scale model parameters to the central server, simultaneously preserving the raw data on local clients. Despite the growing popularity of FD, there is a noticeable gap in previous works concerning the exploration of poisoning attacks within this framework. This can lead to a scant understanding of the vulnerabilities to potential adversarial actions. To this end, we introduce FDLA, a poisoning attack method tailored for FD. FDLA manipulates logit communications in FD, aiming to significantly degrade model performance on clients through misleading the discrimination of private samples. Through extensive simulation experiments across a variety of datasets, attack scenarios, and FD configurations, we demonstrate that LPA effectively compromises client model accuracy, outperforming established baseline algorithms in this regard. Our findings underscore the critical need for robust defense mechanisms in FD settings to mitigate such adversarial threats.

\keywords{Federated Learning  \and Knowledge Distillation \and Knowledge Transfer  \and Poisoning Attack \and Misleading Attack}

\end{abstract}
\section{Introduction}

Federated Learning (FL) \cite{yang2019federated} is an emerging machine learning paradigm that advocates for training a global model in a privacy-preserving manner with local data and computational capabilities over distributed clients fully exploited. Although FL possesses significant advantages over traditional machine learning methods, it faces a series of challenges represented by low communication efficiency during the training process \cite{kairouz2021advances}. Federated Distillation (FD) \cite{jeong2018communication,wu2023improving,itahara2021distill}, a variant of FL, can tackle these issues by only exchanging model outputs (knowledge) during training, without relying on cumbersome model parameters transfer. Moreover, a separate line of works \cite{li2019fedmd,wu2023fedict} demonstrate the superiority of FD over heterogeneous models on clients, which further substantiates the significant role of FD in the FL community.

Despite the notable success of FD, it suffers from security vulnerabilities in practical scenarios. During the training process of FD, the central server is unable to directly monitor the generation of local knowledge, opening the door for malicious participants to disrupt the model training process. Specifically, logits poisoning attacks (LPA) can be an influential methodology to degrade model performance by changing logits extracted on clients, posing significant threats to FD systems. Although poisoning attacks have been sufficiently studied in recent years \cite{sun2021data,yang2023clean,gong2022backdoor,cao2022mpaf,10287523}, there remains a lack of research focusing on the unique security challenges related to LPA that manifest during the FD process.

In this paper, we introduce FDLA, a novel logits poisoning attack method specifically designed for FD. FDLA strategically modifies the transmitted logits sent to the server to mislead the optimization process of local client models. \textcolor{black}{Our proposed manipulation aims to align model outputs with false classification results by generating fake logits that exhibit higher confidence over classes that are similar to the ground truth.} The ultimate objective is to execute attacks that are characterized by precision and effectiveness. This method not only misleads the model into producing incorrect predictions but also injects a high level of credibility into the erroneous predictions. The proposed FDLA significantly adjusts the transfer of logits during the FD process, resulting in a subtle yet effective intervention in the model training of FD. Notably, our attack methodology is executed without necessitating any alterations to private data, thereby charting a new course for interference in the model training procedure.

In general, our contributions can be summarized as follows:
\begin{itemize}
    \item
     \textbf{To our best knowledge, this paper is the first work to investigate logits poisoning attacks in federated distillation.} Our study highlights the security perils associated with knowledge transfer during the training phase, providing a deeper understanding of inherent vulnerabilities in FD.
    \item 
    We propose FDLA, a novel methodology that facilitates logits poisoning attacks in FD. By manipulating the confidence hierarchy of logits, FDLA deceives the model into selecting incorrect, yet seemingly plausible outcomes with increased frequency.
    \item 
	We evaluate our methods on CIFAR-10 and SVHN datasets with different FD settings. Results demonstrate that FDLA achieves more significant model accuracy degression compared with baseline algorithms.
\end{itemize}

\section{Related Work}
\subsection{Poisoning Attack in Federated Learning}

Significant research has been conducted in recent years on poisoning attack strategies that compromise the performance of federated learning models. These strategies include implicit gradient manipulation to attack sample data \cite{sun2021data}, unlabeled attacks \cite{yang2023clean}, backdoor attacks by replacing components of the model \cite{gong2022backdoor}, and the introduction of fake clients to distort the learning process of the global model \cite{cao2022mpaf}. Additionally, there is a technique called VagueGAN that utilizes GAN models to generate blurry data for training toxic local models, which are then uploaded to servers to degrade the performance of other local models \cite{10287523}. These attack methods remain effective against traditional defense mechanisms and norm-clipping strategies. However, these studies primarily focus on conventional federated learning methods and do not consider FD.

\subsection{Federated Distillation}
FD, as a novel paradigm in federated learning that incorporates knowledge distillation \cite{hinton2015distilling,wu2021spirit} for model optimization, has attracted increasing attention in recent years \cite{mora2022knowledge,wu2023survey}. Specifically, FD effectively overcomes system heterogeneity by performing client-side collaborative optimization with knowledge sharing \cite{li2019fedmd,wu2022exploring}; meets the personalized requirements of clients by directing local models to differentiated learning objectives \cite{zhang2021parameterized,wu2023fedict}; reduces the communication bandwidth required for training by transmitting only lightweight model outputs \cite{wu2023fedcache,sattler2021cfd}; enabling training larger models on server via model-agnostic interaction protocols \cite{wu2023agglomerative,cheng2021fedgems}. However, existing methods only focus on the performance improvement of FD algorithms and lack exploration of potential security threats.

\section{Approach}

\subsection{General Process of Federated Distillation}
In the context of FD, each client $k \in \{1,2,...,K\}$ optimizes its model parameters with a combination of cross-entropy loss \(L_{CE}(
\cdot)\) over local data \(D_{k}\) and a distillation loss \( L_{KD}\) over global knowledge \(ZS\) provided by the server. The server orchestrates the collaborative learning process by aggregating the logits (softmax output) from all clients. This iterative training procedure executes numerous communication rounds, with each round conducting a phase of local optimization followed by a phase of global aggregation, culminating in the subsequent update of local model parameters.

Regarding the local optimization phase, each client $k$ optimizes the following objectives:\\
\begin{equation}
\label{J_{k}}
    \mathop {\min }\limits_{{{\rm{W}}_k}} \mathop {J_k}\limits_{(X_k^i,y_k^i)\sim{D_k}} [{J_{CE}} + \beta  \cdot {J_{KD}}],
\end{equation}	
where \( \beta \) is the distillation weighting factor, and \( (X_{k}, y_{k}) \) corresponds to the samples and their associated labels of the local dataset \( D_{k} \). The optimization components \( J_{KD}\) and \( J_{CE}\) can be expressed using the following equations:
\
\begin{equation}
    J_{CE} =  L_{CE}(f(W_{k};X_{k}),y_{k}),
\end{equation}
\begin{equation}
    J_{KD} =  L_{KD}(f(W_{k};X_{k}),ZS),
\end{equation}
where \( W_k \) denotes the model parameters of client $k$, and $f(W_k;\cdot)$ is the prediction function on client $k$.
The server performs the aggregation operation over clients' extracted knowledge \(Z_{k}\), and can be typically formulated as knowledge averaging:
	\begin{equation}\label{aggreation_logits}
	A(Z_{k})=\frac{1}{K}\sum_{k \in \{1,2,...,K\}}Z_{k}
\end{equation}
Upon acquiring global knowledge on the server, it is disseminated to all clients. In response, each client proceeds to update their local models with learning rate $lr$ to align with this shared knowledge, with the process formulated as follows:

\begin{equation}\label{update}
	W_{k} = W_{k} - lr \cdot \nabla_{W_{k}} J_{k}(W_{k}).
\end{equation}

\begin{figure}[h]
	\centering
	\includegraphics[width=\textwidth]{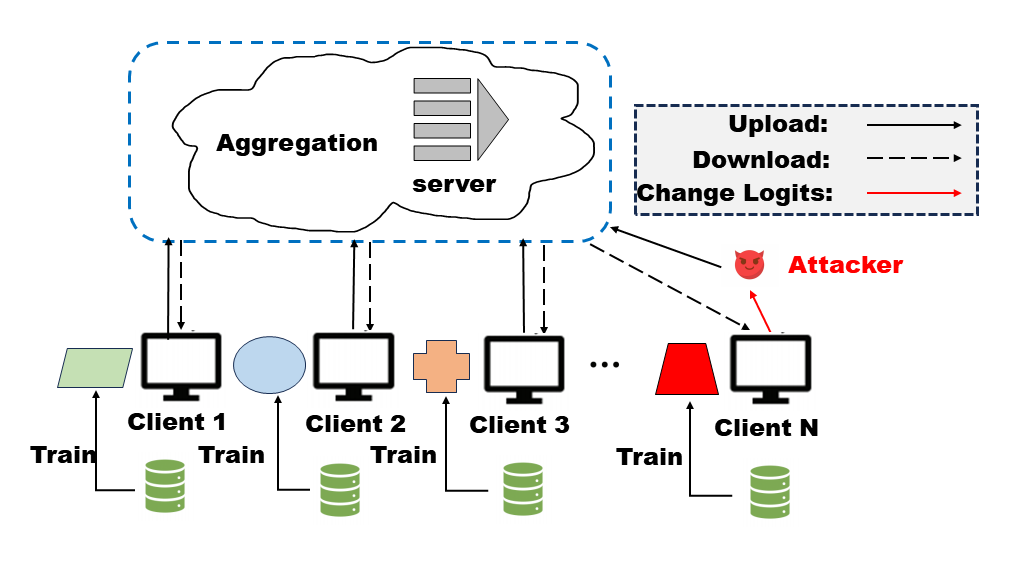}
	\caption{Illustration of how attackers manipulates uploaded knowledge.}
    \label{Logits-Misdirection-Attack-Diagram}
\end{figure}
\FloatBarrier

\subsection{System Model}
FDLA is a malicious method tailored for FD, where a malevolent client disrupts the joint learning process of the model by transforming the confidence of the logits output on local data. "Confidence" represents the model's certainty over its predictions, which measures the reliability of the model's predictions for a given input. When an attacker modifies the confidence values, it can lead to decision biases and hence impact the final outcome. Moreover, since the training process of federated distillation heavily relies on the output knowledge of models, where confidence serves as a crucial component of knowledge, modifying the distribution of confidence can yield changes in the final output of the model. If the rearrangement of confidence values is purposefully manipulated, it can result in a performance decline of the model.Specifically, the client manipulates their knowledge \(Z_{k}\) before uploading them to the server. This manipulation involves sorting and transforming the logits values \(C\) for each sample, where \(C = [c_{1}, c_{2}, ..., c_{n}]\) contains the confidence levels corresponding to each label and \( c_{i} \) represents the confidence of the \( i \)-th element, and assume \( n \) is the total number of corresponding labels.

The transformation process of FDLA is illustrated in Fig. \ref{Logits-Misdirection-Attack-Diagram}: first, the confidences \(c\) are sorted to obtain the sorted indices \(I\). Then, a transformation mapping \(t\) is defined, which maps the element with the highest confidence to the position of the lowest confidence, while the remaining elements are each moved up one position in rank. When the manipulated knowledge \(Z_{k}\) of the malevolent client is uploaded to the aggregation server, the server performs an aggregation operation \(A\). The aggregated global knowledge is then distributed to each client, and the clients update their local models based on this information. However, since the logits stored on the server have been contaminated post-aggregation, the logits finally sent to each client are affected, leading to misdirection in the entire training process.



\subsection{Implementation}
During the \(t\)-th upload of knowledge on the local data \(D_{k}\), the malicious client \(k\) carries out a confidence transformation on the knowledge generated for the \(i\)-th sample \(X_{k}^{i}\) in the current communication round, denoted as \(Z_{k}\). 
We define a confidence ranking function \( r(·) \) and a confidence transformation function \( t(·) \). The ranking function \( r(c_{i}) \) will map \( c_{i} \) to its rank in the confidence vector (with the highest confidence ranked as 1, the second highest as 2, and so on). Then, the transformation function \( t(c_{i}) \) will map \( c_{i} \) to a new confidence \( c_{j} \), \( j \) being the next rank after \( i \), except that the highest confidence is mapped to the lowest confidence. This transformation allows for the identification of the originally highest-confidence label as the least probable, while predicting the most similar results as the correct outcome, and increasing the probabilities of other incorrect labels to further reduce recognition accuracy.

Our proposed execution process is formulated into three steps:
\begin{itemize}
	\item
    [{(\it 1)}]{\it Sort the confidences \(c\) to obtain the sorted indices \(I\).}
	\item
    [{(\it 2)}]{\it Define a transformation mapping \(t\), where \(t[I[1]] = I[n]\), and for \(2 \leq k \leq n\), \(t[I[k]] = I[k - 1]\), where \(I[k]\) is the index of the \(k\)-th highest confidence.}
	\item
    [{(\it 3)}]{\it Apply the transformation mapping \(t\) to the original confidence vector \(c\) to obtain the transformed vector \(c'\).}
\end{itemize}

We express the operation \( CL(c) \) as follows:
\begin{equation}\label{change logits}
	c^{\prime}_{i}=\left\{
	\begin{array}{ll}
		c_{t[i]} &,  \text{if } r(c_{i}) > 1 \\
		c_{t[i]} &,  \text{if } r(c_{i}) = 1
	\end{array} \right.,
\end{equation}
where \( c^{\prime}_{i} \) is the \( i \)-th element in the transformed confidence vector \( c^{\prime} \), \( r(c_{i}) \) is the rank of \( c_{i} \) in \( c \), and \( t[i] \) is the transformed index.
The outcome, which originally should have been identified as correct from \( c_{1} \), is deemed the least likely, while the second most likely result \( c_{2} \) is firmly considered the correct one. When the malicious attacker's \( Z_{k} \) is uploaded to the aggregation server, the server performs an aggregation operation \( A \), as shown in Algorithm \ref{aggregation on server}.

Upon obtaining the aggregated global knowledge, it is distributed to each client. Subsequently, the clients proceed to update their local models in alignment with this newly acquired information.as shown in Algorithm \ref{logits changed on Client k}. During this process, since the logits stored on the server have been contaminated after aggregation, the logits finally sent to each client are affected, which leads to the entire training process being misled.

\begin{algorithm}[t]
	\renewcommand{\algorithmicrequire}{\textbf{Input:}}
	\renewcommand{\algorithmicensure}{\textbf{Output:}}
	\caption{FDLA on Client k}
	\label{logits changed on Client k}
	\begin{algorithmic}[1] 
		\REPEAT
		{
			\FOR{client k in K}
			{
					\STATE $W_{k} \leftarrow W_{k} - lr · \nabla_{wk} · J_{k}(W_{k})$; 	according to Eq.(\ref{update})\\
					\STATE $Z_{k}$ $\leftarrow$ $f(W_{k},X_{k})$;\\
					\IF{client k is venomous client}
					\STATE $Z_{k}$ $\leftarrow$ CL($X_{k}$);   according to Eq.(\ref{change logits})\\
					\ENDIF
					\STATE Upload $Z_{k}$ to the server;\\
			}
			\ENDFOR			
		}				
		\UNTIL Training stop;
	\end{algorithmic}
\end{algorithm}

\begin{algorithm}[t]
	\renewcommand{\algorithmicrequire}{\textbf{Input:}}
	\renewcommand{\algorithmicensure}{\textbf{Output:}}
	\caption{FDLA on the Server}
	\label{aggregation on server}
	\begin{algorithmic}[1] 
		\REPEAT
		\REPEAT
		\STATE Receive $Z_{k}$				
		\UNTIL  Receive all $Z_{k}$ from K clients;
		\STATE // aggregation process
		\STATE ZS $\leftarrow$ A($Z_{k}$) according to Eq(\ref{aggreation_logits});
		\STATE distribute SZ to all clients;
		\UNTIL Training stop;
	\end{algorithmic}
\end{algorithm}

\section{Experiments}
\subsection{Experimental Setup}
We conduct experiments on two common datasets, CIFAR-10 \cite{krizhevsky2009learning} and SVHN \cite{netzer2011reading}, dividing each dataset into 50 not independently and identically distributed parts with the hyperparameter \( \alpha \) that controls the degree of data heterogeneity to 1.0 \cite{wu2023fedcache}. Throughout 200 communication rounds, we ensure convergence for all models. We evaluate our method over two FD frameworks, FedCache \cite{wu2023fedcache} and FD \cite{jeong2018communication}. We adopt the $A^C_1$ in \cite{wu2023fedcache} as the model architectures of clients. Other hyperparameters keep consistent with that adopted in \cite{wu2023fedcache}.


\subsection{Baselines}
We compare our proposed FDLA with two baseline algorithms that modify logits during training: random poisoning, which conducts random modification of logits to make their values random numbers between 0 and 1; and zero poisoning, which sets all logits to zero before uploading.



\begin{table}[h]
\centering
\caption{Average classification accuracy (\%) of models on SVHN dataset using FD and FedCache with varying poisoning ratios, with the \textbf{lowest}  and \uline{second lowest}  accuracies marked.}
\renewcommand\arraystretch{1.3}
\label{main_experiment_1}
\begin{tabular}{c|cccc|cccc}
\hline
\multirow{2}{*}{Poisoning Methods} & \multicolumn{4}{c|}{FD} & \multicolumn{4}{c}{FedCache} \\ \cline{2-9} 
 & \multicolumn{1}{c|}{10\%} & \multicolumn{1}{c|}{20\%} & \multicolumn{1}{c|}{30\%} & Avg. Acc. & \multicolumn{1}{c|}{10\%} & \multicolumn{1}{c|}{20\%} & \multicolumn{1}{c|}{30\%} & Avg. Acc. \\ \hline
No Poisoning & \multicolumn{4}{c|}{78.71} & \multicolumn{4}{c}{46.35} \\ \hline
Random Poisoning & \multicolumn{1}{c|}{\textbf{70.81}} & \multicolumn{1}{c|}{{\ul 65.64}} & \multicolumn{1}{c|}{59.90} & 65.45 & \multicolumn{1}{c|}{{\ul 41.94}} & \multicolumn{1}{c|}{{\ul 39.29}} & \multicolumn{1}{c|}{36.91} & {\ul 39.38} \\ \hline
Zero Poisoning & \multicolumn{1}{c|}{72.35} & \multicolumn{1}{c|}{65.78} & \multicolumn{1}{c|}{\textbf{55.36}} & \textbf{64.50} & \multicolumn{1}{c|}{42.90} & \multicolumn{1}{c|}{39.89} & \multicolumn{1}{c|}{{\ul 36.33}} & 39.70 \\ \hline
FDLA & \multicolumn{1}{c|}{{\ul 71.11}} & \multicolumn{1}{c|}{\textbf{65.30}} & \multicolumn{1}{c|}{{\ul 58.20}} & {\ul 64.87} & \multicolumn{1}{c|}{\textbf{41.23}} & \multicolumn{1}{c|}{\textbf{38.53}} & \multicolumn{1}{c|}{\textbf{34.43}} & \textbf{38.06} \\ \hline
\end{tabular}%
\end{table}

\begin{table}[h]
\centering
\caption{Impact of poisoning on CIFAR-10 model accuracy (\%) using FD and FedCache, with \textbf{lowest} and \uline{second-lowest} accuracies marked.}
\renewcommand\arraystretch{1.3}
\label{main_experiment_2}
\begin{tabular}{c|cccc|cccc}
\hline
\multirow{2}{*}{Poisoning Methods} & \multicolumn{4}{c|}{FD} & \multicolumn{4}{c}{FedCache} \\ \cline{2-9} 
 & \multicolumn{1}{c|}{10\%} & \multicolumn{1}{c|}{20\%} & \multicolumn{1}{c|}{30\%} & Avg. Acc. & \multicolumn{1}{c|}{10\%} & \multicolumn{1}{c|}{20\%} & \multicolumn{1}{c|}{30\%} & Avg. Acc. \\ \hline
No Poisoning & \multicolumn{4}{c|}{52.53} & \multicolumn{4}{c}{52.87} \\ \hline
Random Poisoning & \multicolumn{1}{c|}{\textbf{48.99}} & \multicolumn{1}{c|}{{\ul 44.94}} & \multicolumn{1}{c|}{{\ul 40.72}} & {\ul 44.88} & \multicolumn{1}{c|}{\textbf{47.31}} & \multicolumn{1}{c|}{{\ul 43.46}} & \multicolumn{1}{c|}{{\ul 39.29}} & {\ul 43.35} \\ \hline
Zero Poisoning & \multicolumn{1}{c|}{50.56} & \multicolumn{1}{c|}{45.84} & \multicolumn{1}{c|}{41.20} & 45.87 & \multicolumn{1}{c|}{48.72} & \multicolumn{1}{c|}{45.12} & \multicolumn{1}{c|}{40.25} & 44.70 \\ \hline
FDLA & \multicolumn{1}{c|}{{\ul 49.04}} & \multicolumn{1}{c|}{\textbf{44.82}} & \multicolumn{1}{c|}{\textbf{39.60}} & \textbf{44.49} & \multicolumn{1}{c|}{{\ul 47.39}} & \multicolumn{1}{c|}{\textbf{43.03}} & \multicolumn{1}{c|}{\textbf{38.60}} & \textbf{43.01} \\ \hline
\end{tabular}%
\end{table}
\FloatBarrier

\subsection{Results}
Based on the setup of the main experiment, the final results obtained are shown in Tables \ref{main_experiment_1} and \ref{main_experiment_2}. From the information in the charts, it can be seen that under FDLA (Federated Deep Learning Algorithm), both FD and FedCache are the most effective at impacting model accuracy on the SVHN and CIFAR-10 datasets in the majority of scenarios. However, within the FD algorithm, facing 30\% of malicious attackers, the zero poisoning proved to be the best, achieving the lowest recognition accuracy of 55.35\%. FDLA can reduce the accuracy by up to approximately 20\% in FD and 14\% in FedCache respectively. Additionally, as the number of malicious attackers increases, the impact on model prediction also becomes greater. For instance, in the FD algorithm using the SVHN dataset, as the poisoning ratio increased from 10\% to 30\%, FDLA reduced the average accuracy of each model from about 71\% to 58\%, and the other three experiments showed similar trends. In general, considering all average accuracies, FDLA is the most effective overall.

\begin{figure}[t]
	\subfigure
	{
		\begin{minipage}[b]{0.5\linewidth}
			\centering
			\includegraphics[width=\textwidth]{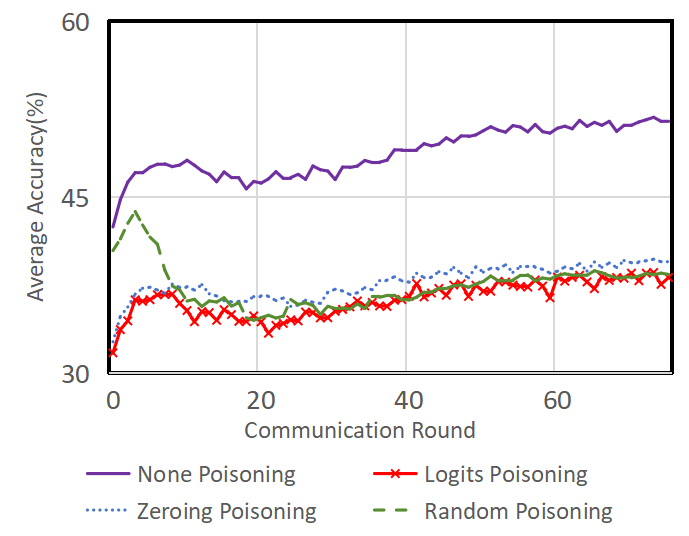}
			\label{impact on convenience in FedCache}
		\end{minipage}
	}
	\subfigure
	{
		\begin{minipage}[b]{0.5\linewidth}
			\centering
			\includegraphics[width=\textwidth]{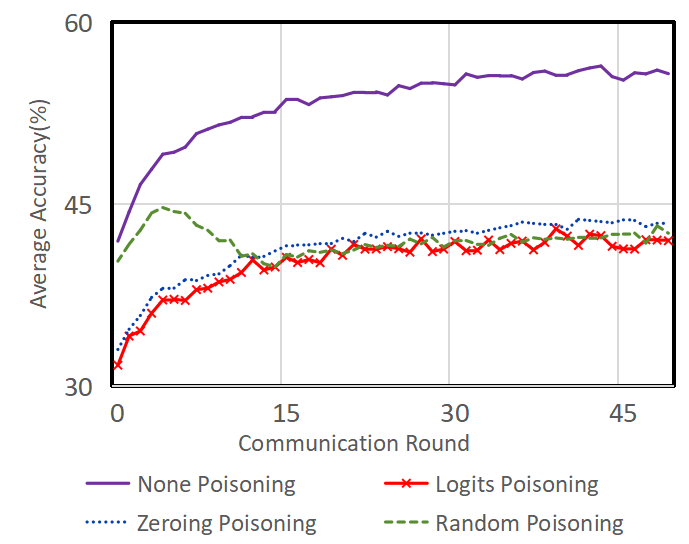}
			\label{impact on convenience in FD}
		\end{minipage}
	}	
	\caption{Convergence impact of three attack types on FD and FedCache in CIFAR-10 with 30\% attackers, tracking average accuracy per communication round.}
	\label{impact on convenience}
\end{figure}

\FloatBarrier

We selected the FD and FedCache algorithms trained on the CIFAR-10 dataset and observed the impact of three different poisoning methods on the model convergence process in a setting with 50 clients and a poisoning ratio of 30\%. As can be seen in Fig. \ref{impact on convenience}, the three methods converge nearly at the same time in both FD and FedCache, around the 50th and 60th rounds of communication, which is consistent with the timing of convergence when there is no poisoning. However, the random poisoning attack causes the model's accuracy to increase slightly at the beginning and then begin to decline before finally converging. The convergence curves of the FDLA and the zero poisoning are highly similar to the convergence curve without poisoning, indicating that these two types of attacks do not affect the model's convergence.

\begin{figure}[t]
	\subfigure
	{
		\begin{minipage}[b]{0.5\linewidth}
			\centering
			\includegraphics[width=\textwidth]{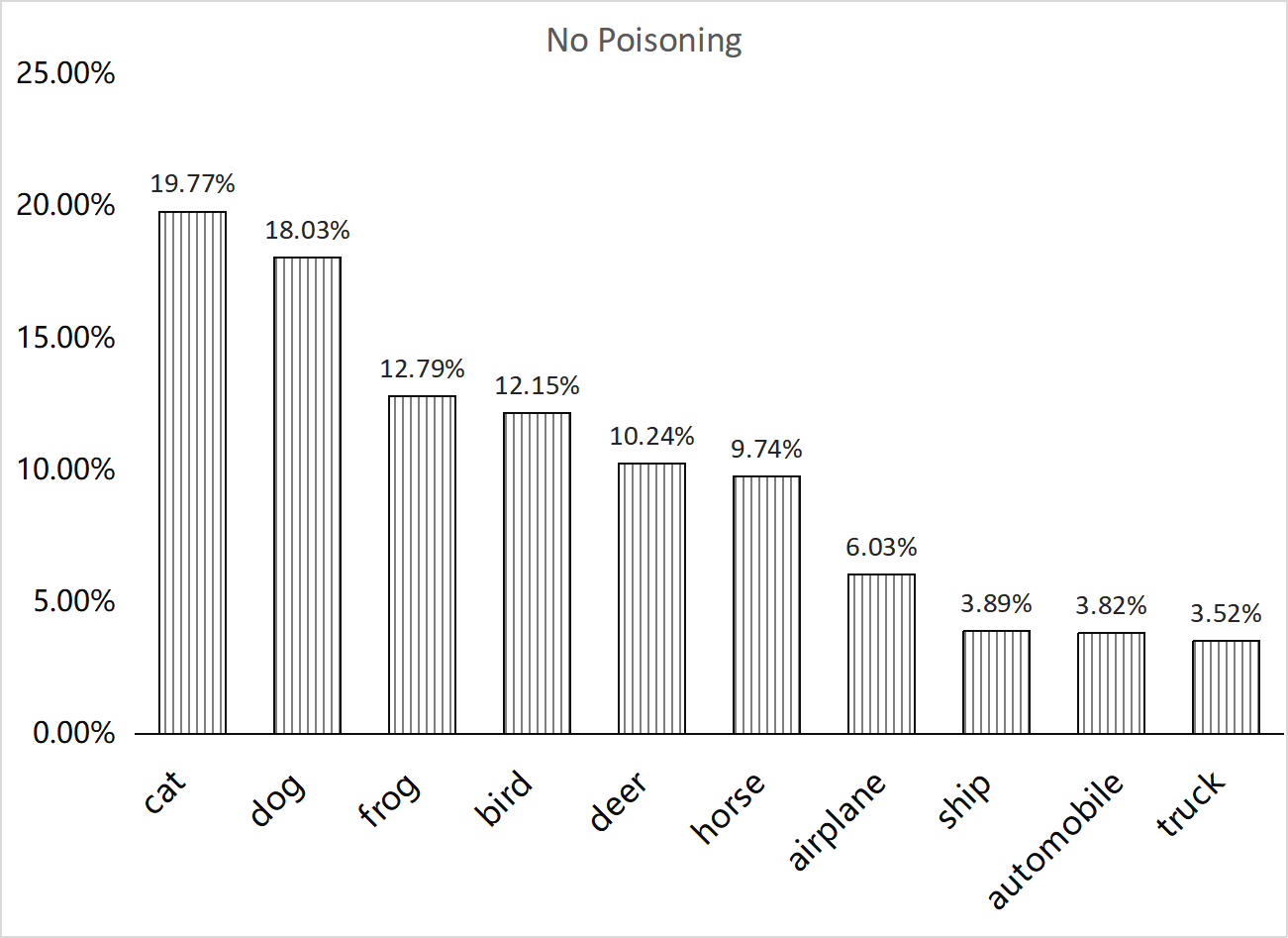}
			\label{The misleading nature of FDLA with no poisioning}
		\end{minipage}
	}
	\subfigure
	{
		\begin{minipage}[b]{0.5\linewidth}
			\centering
			\includegraphics[width=\textwidth]{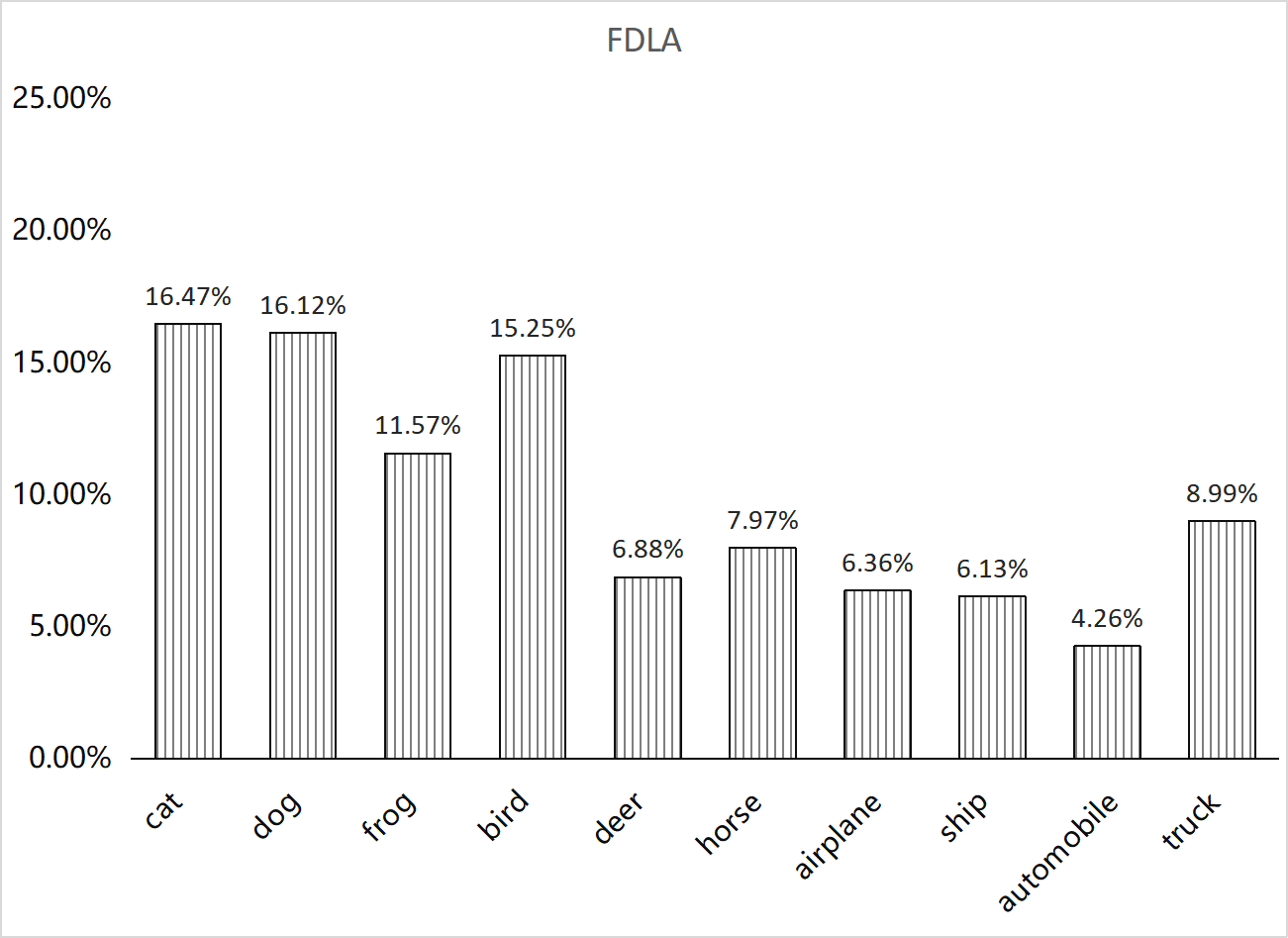}
			\label{The misleading nature of FDLA with FDLA}
		\end{minipage}
	}	
	\caption{Illustration of misleading effect of FDLA to client models. The vertical axis of the chart represents the probability statistics of the test results for cat samples, while the horizontal axis represents the categories of the test results. The left figure shows the predictive statistics of all models on the test set without tampering, while the right figure displays the statistical results after FDLA manipulation.}
	\label{The misleading nature of FDLA}
\end{figure}

\FloatBarrier

To demonstrate the misleading nature of FDLA, we constructed a new dataset by extracting all cat samples from the test set of the CIFAR-10 dataset, along with the training set of CIFAR-10. We employed the FedCache algorithm for 60 rounds of training and testing on 50 clients, as previous studies Fig. \ref{impact on convenience} have indicated that models generally converge around 60 rounds. By statistically analyzing the prediction results of all models on each sample, we evaluated the misleading impact of FDLA.The results, as depicted in Fig. \ref{The misleading nature of FDLA}, revealed that in the absence of tampering, instances where the label "dog" was incorrectly classified as the answer, due to its high similarity to cats, were second only to the correct label "cat". The probability of a cat being predicted was 1.09 times higher than that of a dog. However, after applying FDLA attacks, the probability of correctly identifying cats decreased, while the probabilities of other incorrect labels, such as "car" and "ship", increased. This means that the interference to the models' judgment is successful. Moreover, the ratio of cat probability to dog probability further decreased to 1.02. This provides evidence of FDLA's interference and misleading impact on model predictions.


\subsection{Ablation Study}
The ablation study is divided into three parts, aiming to explore the relationship between hyperparameters and performance:
The default configuration of the experiment is the same as the main experiment, but only the SVHN dataset is used. The basic FD algorithm used is FedCache, with a poisoning ratio of 20\%.
\begin{itemize}
	\item
	The first part investigates the robustness of FDLA concerning data distribution:\\
	The experiment compares the effects of three different poisoning methods by setting $\alpha$ to three distinct values: 0.5, 1.0, and 1.5. This allows for an exploration of how the FDLA's effectiveness is influenced by the distribution of data among the clients.

	\item
	The second part examines the robustness of FDLA concerning the number of clients:\\
	The experiment sets up scenarios with different numbers of participants, specifically 20, 50, and 200, to compare the effects of three distinct poisoning methods. This investigation aims to understand how the effectiveness of FDLA is affected by the number of clients in the federated learning network.

	\item
	The third part of the study explores the robustness of FDLA about model architecture:
	The experiment assesses the performance of three different poisoning methods with models configured to be either homogeneous or heterogeneous, to further investigate the robustness of FDLA to the model structure. The experiment takes into account both the homogeneity and heterogeneity of client models. Specifically, for the homogeneous model experiment, all clients use the same model architecture, denoted as \( A_{1}^{C} \). In the case of heterogeneous models, FedCache assigns different model architectures to clients based on their index modulo 3. Clients with indices that give a remainder of 0, 1, and 2 when divided by 3 are allocated model architectures \( A_{1}^{C} \), \( A_{2}^{C} \), and \( A_{3}^{C} \) respectively. That is :
	\begin{equation}\label{distribute_model}
		A_i^{C} = A_{(i \bmod 3) + 1}^{C}
	\end{equation}

\end{itemize}

\begin{table}[h]
	\centering
	\caption{Poisoning impact on model accuracy (\%) over SVHN dataset with FedCache at 20\% attack ratio and varied data heterogeneity, indicating \textbf{lowest} and \uline{second-lowest} accuracies.}
 \renewcommand\arraystretch{1.3}
 \setlength{\tabcolsep}{10pt}
	\label{different_partition_alpha}
		\begin{tabular}{l|ccc}
			\hline
			Poisoning Methods & $\alpha$=0.5 & $\alpha$=1.0 & $\alpha$=3.0 \\ \hline
			No Poisoning  & 56.67             & 46.35              & 38.04             \\
			Random Poisoning & 46.63             & 39.29             & 32.73             \\
			Zero Poisoning & {\ul 48.78}       & {\ul 39.89}       & {\ul 30.82}       \\
			FDLA & \textbf{46.52}    & \textbf{38.53}    & \textbf{29.77}    \\ \hline
		\end{tabular}
\end{table}

\begin{table}[h]
	\centering
	\caption{Effect of number of clients on accuracy (\%) over SVHN dataset with FedCache over 20\% poisoning, marking \textbf{lowest} and \uline{second-lowest} accuracies.}
 \renewcommand\arraystretch{1.3}
 \setlength{\tabcolsep}{10pt}
	\label{different_clent_number}
		\begin{tabular}{l|ccc}
			\hline
			Poisoning Methods & 20 Clients & 50 Clients & 200 Clients \\ \hline
			No Poisoning & 60.05 & 46.35 & 34.04 \\
			Random Poisoning & {\ul 50.46} & {\ul 39.29} & 30.18 \\
			Zero Poisoning & 51.12 & 39.89 & {\ul 28.47} \\
			FDLA & \textbf{47.32} & \textbf{38.53} & \textbf{28.17} \\ \hline
		\end{tabular}%
\end{table}

\begin{table}[h]
	\centering
	\caption{Impact of model structures on accuracy (\%) over SVHN dataset with FedCache over 20\% poisoning, highlighting \textbf{lowest} and \uline{second-lowest} accuracies.}
 \renewcommand\arraystretch{1.3}
 \setlength{\tabcolsep}{10pt}
	\label{different_heterogeneity}
		\begin{tabular}{l|cc}
			\hline
			Poisoning Methods & Homogeneous  & Heterogeneous \\ \hline
			No Poisoning & 46.35 & 58.07 \\
			Random Poisoning & {\ul 39.29} & 48.93 \\
			Zero Poisoning & 39.89 & {\ul 48.74} \\
			FDLA & \textbf{38.53} & \textbf{46.63} \\ \hline
		\end{tabular}%
\end{table}
\FloatBarrier

Table \ref{different_partition_alpha} shows the impact of data distribution on FDLA, random poisoning, and zeroing poisoning. The results indicate that as heterogeneity increases, the impact of poisoning on the model's judgment remains around 10\%, with FDLA being the most effective. As the variety of data types in a dataset increases, FDLA maintains almost the same level of deceptiveness, indirectly demonstrating its robustness to data heterogeneity.

Table \ref{different_clent_number} shows the effect of the number of participants on FDLA, random poisoning, and zeroing poisoning. The results suggest that as the number of clients increases, the impact of poisoning on model judgment slightly decreases, with FDLA again being the most effective. However, this diminishing effect may be due to the increased number of clients and the data preprocessing rules for data partitioning, which result in a reduction of local data per client, as well as the algorithms used.

Table \ref{different_heterogeneity} shows the impact of model structure on FDLA, random poisoning, and zeroing poisoning. The results show that regardless of whether the models are heterogeneous or homogeneous, FDLA causes around a 10\% decrease in model judgment accuracy. This suggests that FDLA has a certain degree of robustness to the structure of the model.

\section{Conclusion}
We propose FDLA, which is a logits poisoning attack method for federated distillation. FDLA  modifies the logits received by the server to guide the model to generate highly confident incorrect predictions. It finely tunes the knowledge-sharing mechanism during the federated distillation process, subtly affecting model training. To our best knowledge, this paper is the first work to investigate logits poisoning attacks in federated distillation. Experimental results prove the strong robustness of FDLA when facing diverse model and data settings. This research not only showcases vulnerabilities that attackers might exploit but also provides a new baseline into directions for defending against attacks targeting federated distillation.

\begin{credits}
\subsubsection{\ackname} This work was supported in part by the Fundamental Research Funds for the Central Universities under Grant 2021JBM008 and Grant 2022JBXT001, and in part by the National Natural Science Foundation of China (NSFC) under Grant 61872028. (Corresponding author: Bo Gao.)

\end{credits}


%
%
%
\bibliographystyle{unsrt}

%




\end{document}